\documentclass[final,3p,times]{elsarticle}

\usepackage{amssymb}
\usepackage{hyperref}
\usepackage{graphicx}
\usepackage{multirow}
\usepackage{xcolor}

\journal{Engineering Applications of Artificial Intelligence}

\begin{document}

\begin{frontmatter}



\title{A Culturally-Aware Benchmark for Person Re-Identification in Modest Attire}


\author[inst1]{Alireza Sedighi Moghaddam}
\ead{a\_sedighi77@comp.iust.ac.ir}

\author[inst1]{Fatemeh Anvari}
\ead{fatemeh\_anvari@comp.iust.ac.ir}

\author[inst1]{Mohammadjavad Mirshekari Haghighi}
\ead{mohammadjavad.mirshekari@gmail.com}

\author[inst1]{Mohammadali Fakhari}
\ead{sm\_fakhari@comp.iust.ac.ir}

\author[inst1]{Mohammad Reza Mohammadi\corref{cor1}}
\ead{mrmohammadi@iust.ac.ir}

\affiliation[inst1]{organization={School of Computer Engineering, Iran University of Science and Technology},
            country={Islamic Republic of Iran}}

\cortext[cor1]{Corresponding author}

\begin{abstract}
Person Re-Identification (ReID) is a fundamental task in computer vision with critical applications in surveillance and security. Despite progress in recent years, most existing ReID models often struggle to generalize across diverse cultural contexts, particularly in Islamic regions like Iran, where modest clothing styles are prevalent. Existing datasets predominantly feature Western and East Asian fashion, limiting their applicability in these settings. To address this gap, we introduce Iran University of Science and Technology Person Re-Identification (IUST\_PersonReId), a dataset designed to reflect the unique challenges of ReID in new cultural environments, emphasizing modest attire and diverse scenarios from Iran, including markets, campuses, and mosques.

Experiments on IUST\_PersonReId with state-of-the-art models, such as Semantic Controllable Self-supervised Learning (SOLIDER) and Contrastive Language-Image Pretraining Re-Identification (CLIP-ReID), reveal significant performance drops compared to benchmarks like Market1501 and Multi-Scene MultiTime (MSMT17), specifically, SOLIDER shows a drop of 50.75\% and 23.01\% Mean Average Precision (mAP) compared to Market1501 and MSMT17 respectively, while CLIP-ReID exhibits a drop of 38.09\% and 21.74\% mAP, highlighting the challenges posed by occlusion and limited distinctive features. Sequence-based evaluations show improvements by leveraging temporal context, emphasizing the dataset's potential for advancing culturally sensitive and robust ReID systems. IUST\_PersonReId offers a critical resource for addressing fairness and bias in ReID research globally.


\end{abstract}




\end{frontmatter}




\section{Introduction} \label{sec:intro}  

Person Re-Identification (ReID) is a crucial task in computer vision, aimed at identifying and matching individuals across images captured by different cameras at various times and locations. The primary motivation for ReID arises from its vital applications in surveillance, security, and urban management. Surveillance cameras, often deployed across vast areas, generate massive amounts of footage. Manually sifting through this data to track individuals is impractical. Thus, automated ReID systems provide a scalable and efficient solution for matching identities across non-overlapping camera views, enabling more effective monitoring and management of public spaces.

The performance of Artificial Intelligence (AI) models, heavily depends on the quality and diversity of their training datasets, and ReID models are no exception. Developing robust models requires datasets that capture a wide range of variations in lighting, viewpoints, quality, and human appearances. Among these factors, clothing is a significant determinant of human appearance, varying dramatically between different cultures. For instance, women's clothing in countries where hijabs are commonly worn, such as Iran, differs markedly from clothing in other regions. Incorporating cultural clothing diversity in ReID datasets is essential. These datasets not only improve model accuracy and reliability but also ensure fairness by preventing demographic bias in real-world applications \citep{perkowitz2021bias}. Without sufficient representation, models may perform poorly on underrepresented groups, leading to biased and unfair outcomes.

In recent years, several person ReID datasets have been developed, each addressing unique challenges. The Chinese University of Hong Kong (CUHK03) \citep{li2014deepreid} dataset captures 1,364 pedestrians across six surveillance cameras, introducing cross-view complexities with varying pedestrian movements. The Duke Multi-Target, Multi-Camera Re-Identification (DukeMTMC-reID) \citep{narayan2017person, zheng2017unlabeled} dataset, derived from DukeMTMC, includes 1,852 identities across eight cameras, tackling real-world challenges like occlusions, illumination changes, detection errors, and low-resolution imagery. Market-1501 \citep{zheng2015scalable} provides over 32,000 annotated bounding boxes from six cameras, featuring distractor images and intra-class variations in a supermarket setting. MSMT17 \citep{wei2018person} spans 15 cameras and captures 4,101 identities under varying lighting, weather, and time conditions, testing model robustness in both indoor and outdoor environments.

Unlabeled datasets have gained significant attention for pre-training deep models. These datasets can easily scale, reduce data collection costs, and are particularly valuable for learning robust feature representations. Large-scale Unlabeled Person Re-Identification (LUPerson) \citep{fu2021unsupervised}, for instance, comprises 4 million images from 200,000 identities, predominantly derived from studio-quality videos on YouTube, rather than real-world surveillance conditions. This dataset spans diverse scenes such as streets, campuses, and sports fields, enabling large-scale pre-training for ReID. LUPerson-NL \citep{fu2022large}, an extension of LUPerson with 10 million images and 430,000 identities, introduces noisy labels through automated tracking, offering greater diversity in person representations.

The primary objective of collecting the IUST\_PersonReId dataset is to address the gap in cultural clothing representation between Iran and other regions. This gap leads to a substantial performance drop in ReID models when applied to environments where these specific clothing patterns are prevalent. In these contexts, individuals dress more modestly than is common in many Western regions (such as the USA and Europe). Similarly, the clothing styles differ significantly from those in parts of East Asia (such as China and India). This modest attire, particularly women wearing hijabs, makes re-identification of individuals a more challenging task compared to scenarios represented in existing datasets. 

To tackle these challenges, we present the IUST\_PersonReId dataset, which encompasses unique cultural and environmental conditions. This dataset is designed to fill the existing gap and provide a robust resource for training and evaluating ReID models in contexts where modest clothing is prevalent. We provide an extensive evaluation procedure to validate our dataset and its significance for this domain. We show that the observed performance decrease due to cultural clothing differences is real, and the dataset effectively addresses this issue. Our evaluations demonstrate significant improvements in the new domain, reflected by an increase in rank-1 accuracy by about 40\%. The dataset is publicly available at \textcolor{purple}{\href{https://computervisioniust.github.io/IUST_PersonReId/}{https://computervisioniust.github.io/IUST\_PersonReId/}}.

The structure of the paper is as follows: \hyperref[sec:related]{Section 2} reviews related works. \hyperref[sec:methodology]{Section 3} describes methodology of data collection, annotation, and dataset composition. \hyperref[sec:results]{Section 4} discusses model performance, the impact of cultural clothing, and key insights. Finally, \hyperref[sec:conclusion]{Section 5} provides the conclusions and potential directions for future work.
\section{Related Works}
\label{sec:related}

Person re-identification (ReID) has been a crucial task in computer vision, aiming to identify the same person across different cameras or time. Over the years, numerous datasets have been proposed to facilitate research in this area. This section provides an overview of some of the most popular and influential ReID datasets.

\subsection{Supervised Datasets}

\paragraph{East Asian culture} The CUHK03 dataset \citep{li2014deepreid}, created by the Chinese University of Hong Kong, consists of images and videos captured on the university's campus using two pairs of cameras. The dataset represents East Asian modern clothing culture and includes both manually labeled and automatically detected bounding boxes, with the latter generated by the Deformable Parts Model (DPM) algorithm \citep{felzenszwalb2009object}.
The Market-1501 dataset \citep{zheng2015scalable} is a widely used resource collected at the Tsinghua University campus supermarket with up to six cameras, capturing East Asian modern clothing culture and including detections from the DPM algorithm \citep{felzenszwalb2009object}. An extension of this dataset, Motion Analysis and Re-identification Set (MARS) \citep{zheng2016mars}, was also collected on the Tsinghua University campus using up to six cameras. It encompasses clothing styles same as Market-1501, with detections provided by both the DPM and Generalized Maximum Multi-Clique Problem (GMMCP) \citep{dehghan2015gmmcp} algorithms.

\paragraph{Western and European culture} DukeMTMC4ReID \citep{ristani2016performance} is another widely recognized dataset, captured by up to eight cameras on the Duke University campus. It includes Western modern clothing styles in an academic environment, with detections facilitated by the Doppia \citep{benenson2015ten} algorithm. DukeMTMC-VideoReID \citep{wu2018exploit} is a subset of the DukeMTMC tracking dataset \citep{ristani2016performance} for person re-identification captured by up to eight cameras on the Duke University campus. The Airport dataset \citep{gou2018systematic} was created using videos from six cameras within an indoor surveillance network at a mid-sized airport. It primarily features Western modern clothing styles but also includes a variety of international styles due to the diverse nature of the airport environment. DeepSportradar-ReID \citep{felsen2017will} is a dataset containing 4,869 images of 486 identities, each captured by multiple cameras. The RPIfield dataset \citep{zheng2018rpifield} was created using 12 cameras on the campus of Rensselaer Polytechnic Institute. It captures individuals wearing Western modern clothing in an academic environment, with each identity observed from 12 different camera views. The SoccerNet-ReID dataset \citep{giancola2022soccernet} was created on soccer fields across six European soccer leagues. It focuses on soccer sportswear and introduces the challenge of distinguishing identities wearing similar clothing, as is typical for players on the same team.

\paragraph{Multiple cultures} The Pedestrian Detection, Tracking, and Re-Identification (DESTRE-P) dataset \citep{wu2020destre} was collected in two different countries, India and Portugal, and represents the clothing culture of both regions in an academic environment. Unlike other datasets, it is not limited to a single location or country. The Large-scale Spatio-Temporal Person Re-Identification (LaST) \citep{shu2021large} dataset created by collecting 2k movies, covering more than 8 countries from Asia to Europe. This dataset focus on enlarging spatial scope and time span of pedestrian activities by using movies sources. This dataset covers more diverse clothing culture in comparison of previous datasets by covering multiple countries.

\paragraph{Unspecified Cultural Context} The MSMT17 dataset \citep{wei2018person} was collected on a university campus. However, it lacks specific information regarding the location and the clothing culture represented within the dataset. The Labeled Pedestrian in the Wild (LPW) dataset \citep{song2018region} comprises 2,731 identities, each captured by up to four cameras. However, there is no information available about the location where the dataset was collected. The Multi-Attribute and Language Search (MALS) dataset \citep{yang2023towards} is a recent addition to the field, designed to address the challenges of text-based person retrieval through a large-scale, multi-attribute, and language search benchmark. The dataset is generated to simulate diverse pedestrian appearances, environments, and multi-modal queries. It includes over 1.5 million synthetic images and associated textual descriptions, making it one of the largest datasets in this domain. This dataset does not adhere to the clothing styles of any specific culture or country, as it was entirely generated using synthetic methods.

\subsection{Unsupervised Datasets}
Unsupervised datasets have gained significant attention in recent years due to their ability to scale easily and reduce the costs associated with data collection and labeling. The LUPerson dataset \citep{fu2021unsupervised} comprises over 4 million pedestrian images, estimated to include over 200k identities, sourced from a diverse range of environments, aimed at improving unsupervised ReID models through the provision of extensive unlabeled data. The dataset was curated by crawling over 70,000 street-view videos from YouTube, using search queries such as “city name + street view (or scene)”. To ensure broad diversity, the dataset includes videos from the top 100 largest cities globally, predominantly located in Western countries and East Asia, thereby reflecting a wide array of clothing cultures.
Building upon LUPerson, the LUPerson-NL dataset \citep{fu2022large} introduces noisy labels to simulate real-world labeling errors, which is crucial for the development and evaluation of ReID algorithms capable of handling label noise. Each tracklet generated by the tracking algorithm is considered a distinct identity and referred to as a noisy label. This extension is designed to ensure that ReID models can maintain high performance even when confronted with inaccuracies in the training data. Table \ref{tab:datasets} provides a summary of the datasets mentioned above.

\begin{table*}[h]
\centering
\caption{Summary of Person ReID Datasets}
\label{tab:datasets}
\renewcommand{\arraystretch}{1.5}
\resizebox{\textwidth}{!}{%
\begin{tabular}{|l|c|c|c|c|c|c|}
\hline
\textbf{Dataset} & \textbf{\# Identities} & \textbf{\# Cameras} & \textbf{Country, City} & \textbf{Location} & \textbf{\# Images/Videos} & \textbf{Year} \\ \hline
CUHK03 \citep{li2014deepreid} & 1,467 & 6 & China, Hong Kong & Chinese University of Hong Kong's campus & 13,164 & 2014 \\ \hline
Market-1501 \citep{zheng2015scalable} & 1,501 & 6 & China, Beijing & Tsinghua University campus supermarket & 32,668 & 2015 \\ \hline
MARS \citep{zheng2016mars} & 1,261 & 6 & China, Beijing & Tsinghua University campus & 20,715 & 2016 \\ \hline
DukeMTMC4ReID \citep{ristani2016performance} & 1,852 & 8 & USA, North Carolina, Durham & Duke University campus & 46,261 & 2017 \\ \hline
Airport \citep{gou2018systematic} & 1,382 & 6 & USA, Massachusetts, Boston & Mid-sized Airport & 39,902 & 2017 \\ \hline
DeepSportradar-ReID \citep{felsen2017will} & 486 & Multiple & France & French basketball league & 4,869 & 2017 \\ \hline
DukeMTMC4ReID-Video \citep{wu2018exploit} & 1,852 & 8 & USA, North Carolina, Durham & Duke University campus & 46,261 & 2018 \\ \hline
RPIfield \citep{zheng2018rpifield} & 112 & 12 & USA, New York, Troy & Rensselaer Polytechnic Institute campus & 601,581 & 2018 \\ \hline
MSMT17 \citep{wei2018person} & 4,101 & 15 & - & University campus & 126,441 & 2018 \\ \hline
SoccerNet-ReID \citep{giancola2022soccernet} & 2,933 & 6 & Europe & Six European soccer leagues & 34,093 & 2018 \\ \hline
LPW \citep{song2018region} & 2,731 & 4 & - & - & 7,694 & 2018 \\ \hline
DESTRE-P \citep{wu2020destre} & 1,121 & Multiple & Portugal \& India & Beira Interior \& JSS Science Technology Universities campus & - & 2020 \\ \hline
LUPerson \citep{fu2021unsupervised} & $>200k$ & 1000 & Top 100 big cities & YouTube streetviews & 46,260 & 2021 \\ \hline
LUPerson-NL \citep{fu2022large} & $\simeq$ 433,997 & 21,697 & Top 100 big cities & YouTube streetviews & 10,683,716 & 2022 \\ \hline
LaST \citep{shu2021large} & 10,862 & Multiple & 8+ countries from Asia to Europe & Movies & 228,156 & 2022 \\ \hline
MALS \citep{yang2023towards} & - & Synthesis & Synthesis & Synthesis & 1,510,330 & 2023 \\ \hline
IUST\_PersonReId & 1,847 & 19 & Iran \& Iraq & IUST University campus, Market, Mosque, Streetview & 117,455 & 2024 \\ \hline
\end{tabular}%
}
\end{table*}

Despite these advancements, none of the prior datasets specifically target the unique challenges posed by cultural clothing variations in Islamic countries. In this paper, we introduce the IUST\_PersonReId dataset, which is specifically designed to include images from Islamic regions, addressing the significant cultural and clothing differences. This dataset aims to improve the performance of ReID models in these unique contexts, ensuring more accurate and reliable re-identification results in environments with distinct cultural practices.
\section{Methodology}
\label{sec:methodology}

We followed a clear and structured process that involved collecting data from real-world settings, adding challenging conditions on purpose, carefully labeling the data, and using thoughtful sampling methods. The following subsections detail each of these components.

\subsection{Data Collection}
The raw videos were gathered from five distinct locations: the campus of Iran University of Science \& Technology, a fruit shop, a hypermarket, a mosque, and the Arbaeen procession in Iraq, which is one of the largest gatherings of Muslims in the world. To ensure that the dataset reflects real-world scenarios, surveillance camera footage was primarily used during the data collection process. However, for the Arbaeen procession, handheld cameras were employed to capture videos of the participants. This approach enabled us to gather raw video data that encompasses a wide variety of cultural representations, which are largely absent in existing datasets.

\subsection{Supported Challenges in Data Collection}

To enhance the robustness and real-world applicability of our dataset, we intentionally designed the data collection process to include various challenging scenarios encountered in surveillance systems. These challenges are described below:

\begin{itemize}
    \item \textbf{Special Camera Angles:} The dataset includes videos captured from unique and non-standard camera angles typical of surveillance cameras, ensuring coverage of diverse viewpoints.
    
    \item \textbf{Changing Lighting Conditions:} The dataset simulates transitions between bright outdoor environments and dark indoor spaces, mimicking real-world scenarios where lighting conditions change dynamically as a person moves.
    
    \item \textbf{Varying Camera Quality:} To reflect the diversity in surveillance hardware, the dataset includes videos from both normal and high-quality surveillance cameras.
    
    \item \textbf{Seasonal Clothing Variations:} Our dataset accounts for significant clothing changes across seasons, including lighter summer attire and heavier winter garments.
    
    \item \textbf{Similar Clothing Scenarios:} Videos were recorded during Muharram religious ceremonies and the Arbaeen procession, where the majority of individuals wear black clothing as a symbol of mourning. This poses a unique challenge in distinguishing individuals with similar attire.

\end{itemize}

\begin{figure*}[ht]
    \centering
    \includegraphics[width=\linewidth]{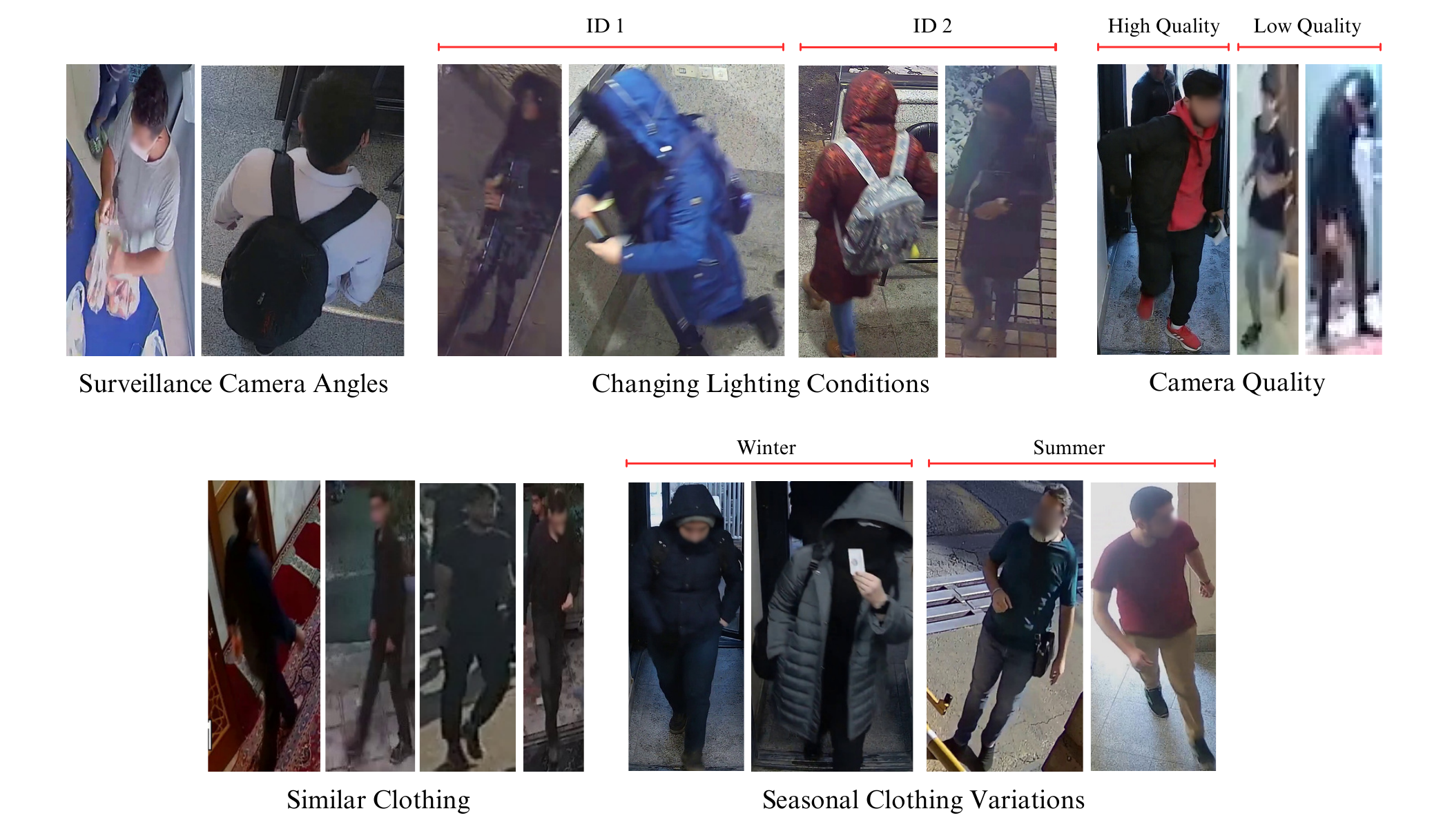}
    \caption{Examples of challenges supported in our dataset, including special camera angles, lighting variations, seasonal clothing and similar clothing.}
    \label{fig:challenges-examples}
\end{figure*}

To illustrate these challenges, Figure~\ref{fig:challenges-examples} presents example images showcasing the variety of scenarios included in the dataset.

\subsection{Annotation Process}
In this phase the raw videos split into 30 minutes partitions. After that we use pedestrian tracking algorithms to pre-annotate each of the partitions. Here we use multiple tracking algorithms suitable for each environment cause each of the algorithms surpasses other algorithms in specific environments and we use them all. We use You Only Look Once version-5 (YOLOv5) \citep{Jocher_YOLOv5_by_Ultralytics_2020}, YOLOv8 \citep{Jocher_YOLOv5_by_Ultralytics_2020} with bytetrack \citep{zhang2022bytetrack} tracker, Fair Multi Object Tracking (FairMOT) \citep{zhang2021fairmot} and YOLOE \citep{ppdet2019} models which have builtin trackers. After auto annotation step, we conduct a filtering procedure to make supervised annotation process much easier. We remove bounding boxes which appeared in less than 5 frames because during the annotation these bounding boxes appear for a very short time and disappear quickly and can't be easily seen by annotator to validate or edit it. We use Computer Vision Labelling Tool (CVAT) \citep{CVAT_ai_Corporation_Computer_Vision_Annotation_2023} application for make corrections in automated annotations and make it available online to all of annotators to make them work remotely. To help them download the frames much faster we conduct resizing the videos with respect to their file size with using compression strategies available in this platform.

Prior to hiring annotators for the annotation process, we conducted a preliminary annotation study. This initial effort was aimed at identifying the challenges inherent in the annotation task, thereby enabling the development of comprehensive guidelines to ensure high-quality annotations. Following this, we provided annotators with detailed documentation and educational videos to clarify the process, which helped minimize annotation errors. We recruited over 20 annotators for the detection and tracking tasks, and implemented a qualification process involving annotation tests to ensure the highest possible accuracy and consistency in the results.

\begin{figure*}[ht]
    \centering
    \includegraphics[width=\linewidth]{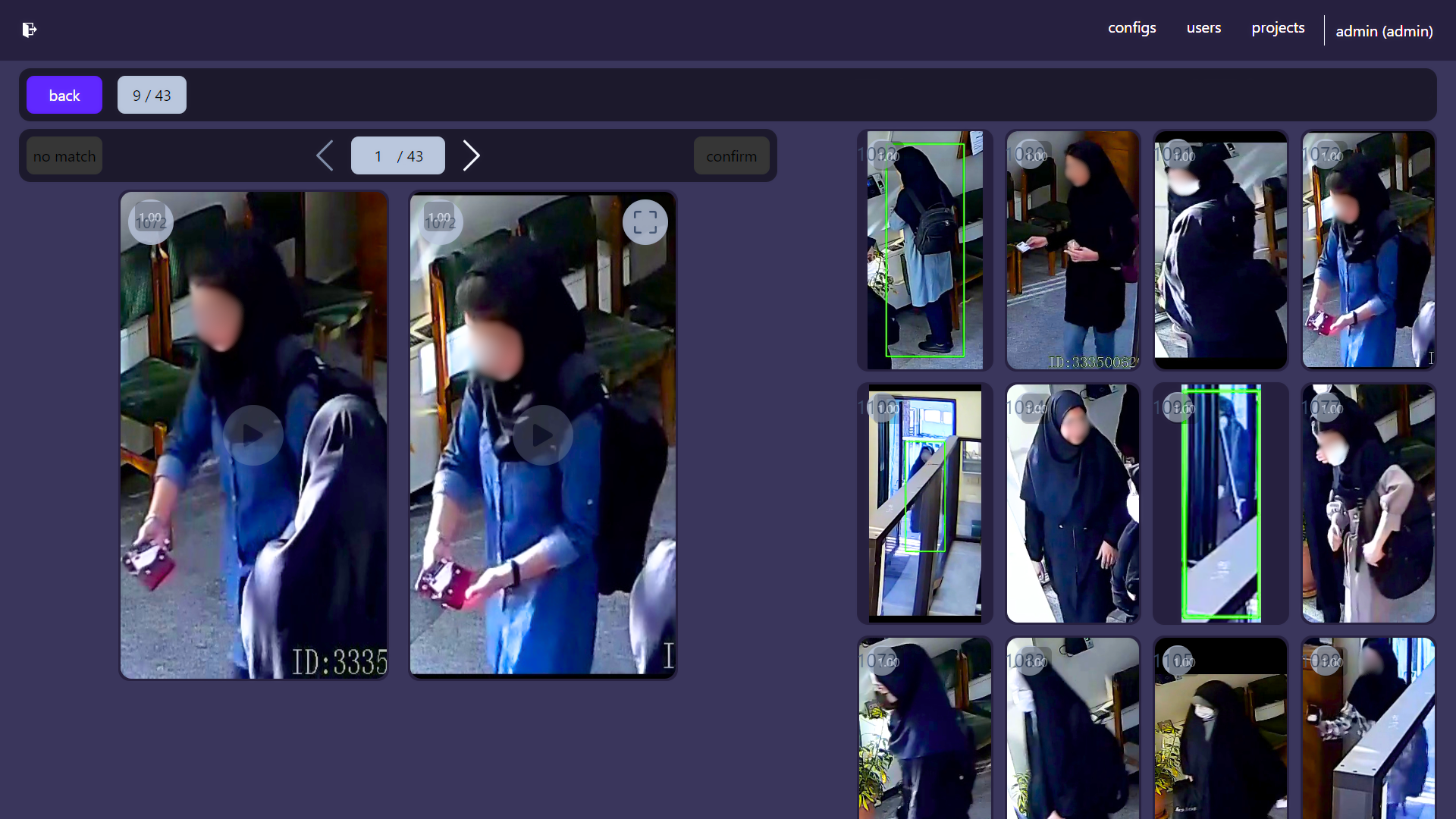}
    \caption{A screenshot of the CVLab-ReID-Tool designed for efficient data annotation in the re-identification phase.}
    \label{fig:cvlab-reid-tool}
\end{figure*}

The generation of supervised data for the re-identification task involves annotating data to match the same individuals across different camera views (cross-camera re-identification) and within the same camera as they exit and re-enter the scene (self re-identification). Although tools such as 
Rapid-Rich Object Search (ROSE) ReID \citep{ROSEReID} and the Semi-Automatic Data Annotation Tool \citep{zhao2018semi} exist, they are not publicly accessible. Consequently, we developed our own annotation tool for the re-identification phase, named Computer Vision Lab Re-Identification Tool (CVLab-ReID-Tool) \footnote{\href{https://github.com/ComputerVisionIUST/CVLab-ReId-Tool}{https://github.com/ComputerVisionIUST/CVLab-ReId-Tool}}, which we have made publicly available. An example of the designed app can be seen in Figure \ref{fig:cvlab-reid-tool}.  

Given the critical nature of this annotation phase, we employed the three most experienced members of our lab to perform the annotations, thereby minimizing annotation noise. To facilitate the annotation process, similar to the detection and tracking phase, we integrated a re-identification model that ranks gallery IDs based on their similarity to the query, simplifying the selection process. Additionally, we incorporated the temporal difference in the appearance of identities as an additional feature for sorting gallery items. By leveraging these two features, we consistently identified the correct match within the top-10 candidates in the sorted gallery list.

\subsection{Sampling Procedure}
Selecting a representative and diverse subset of frames is crucial for creating an efficient and effective dataset while avoiding redundancy and ensuring computational feasibility. To ensure the IUST\_PersonReId dataset provides high-quality, representative frames for each individual, we designed a robust sampling procedure. Rather than including all frames from annotated videos, our approach samples a subset of frames using a two-step process to balance computational efficiency and data diversity.

\paragraph{Frame Sampling}  
Frames are sampled from each annotated video at intervals of 250 ms. Since the raw videos have different Frames Per Second (FPS) rates, this interval corresponds to $0.25 \times \mathrm{FPS}$, giving the frame number interval equivalent to 250 ms. This approach ensures uniform sampling across videos with varying FPS. Additionally, we limit the number of frames for each individual to a maximum of 50, which helps reduce redundancy and dataset size while maintaining sufficient diversity. For individuals with fewer than 50 frames, all available frames are included.

\paragraph{Quality-Based Selection}
For individuals with more than 50 sampled frames, we employ a quality assessment algorithm to select the best-quality frames. The Blind Reference-less Image Spatial Quality Evaluator (BRISQUE) \citep{mittal2012no} algorithm is used to assess the visual quality of each frame. Using a sliding window approach, we evaluate consecutive subsets of frames and select the subset with the highest aggregate quality score. This ensures the dataset prioritizes clear and visually informative frames, which are essential for robust model training and evaluation.

\subsection{Dataset Composition}
\begin{figure}[ht]
    \centering
    \begin{minipage}{0.48\textwidth}
        \centering
        \includegraphics[width=\textwidth]{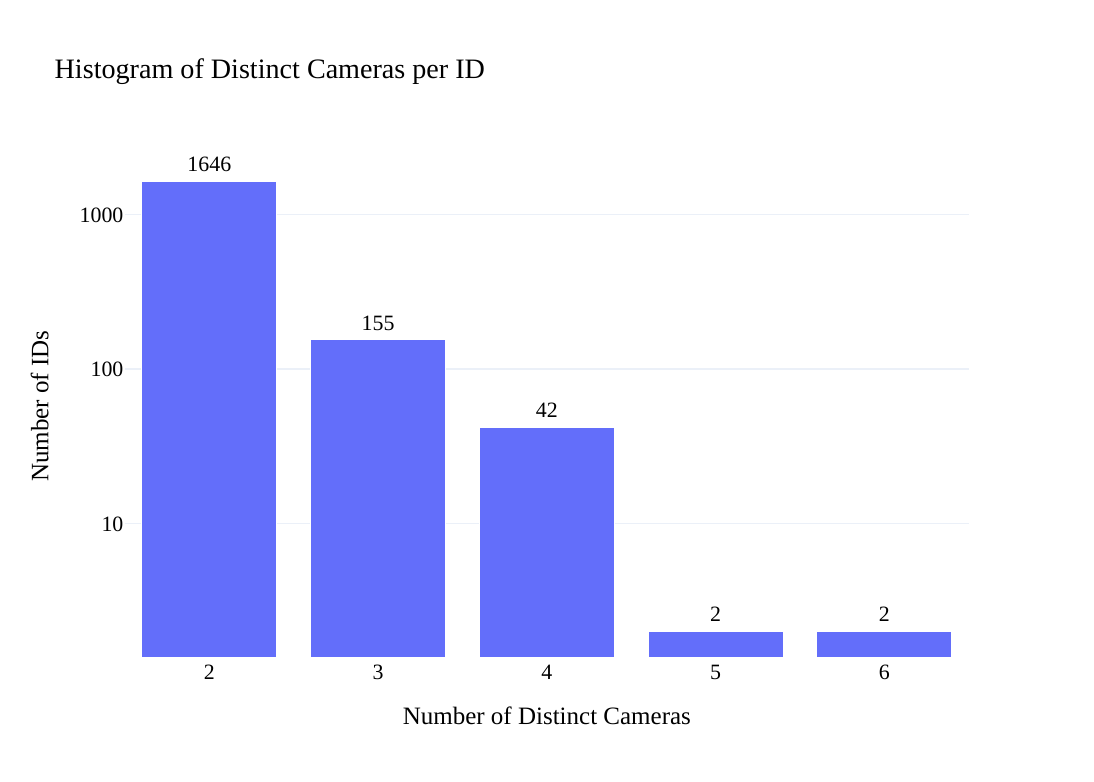}
        \caption{Number of cameras that captured each identity.}
        \label{fig:camera_numbers}
    \end{minipage}
    \hfill 
    \begin{minipage}{0.48\textwidth}
        \centering
        \includegraphics[width=\textwidth]{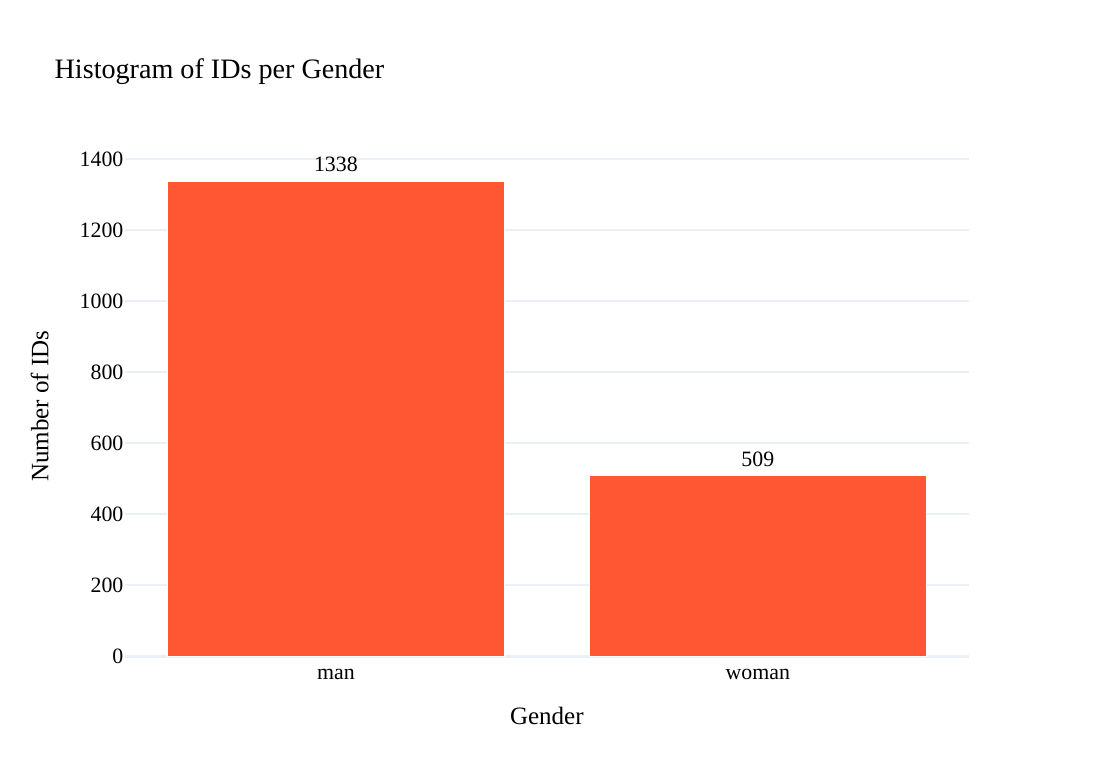}
        \caption{The number of identities for each gender}
        \label{fig:gender_numbers}
    \end{minipage}
\end{figure}

The dataset was constructed by gathering about 500 hours of raw video footage from five distinct environments, encompassing both indoor and outdoor settings. These environments include the campus of Iran University of Science \& Technology, a fruit shop, a hypermarket, a mosque, and a street view of the Arbaeen procession in Iraq.
IUST\_PersonReId consists of 1,847 unique identities, represented by 117,455 annotated bounding boxes. The distribution of captured identities across the number of cameras utilized is depicted in Figure \ref{fig:camera_numbers}. Additionally, the gender distribution within the dataset is visualized in Figure \ref{fig:gender_numbers}.

\subsection{Models and Training}
The SOLIDER \citep{chen2023beyond} model was initialized with weights pre-trained on LUPerson \citep{fu2021unsupervised}, while the CLIP-ReID \citep{li2023clip} model uses the pre-trained CLIP model \citep{radford2021learning} without additional pre-training. Both models were fine-tuned on our dataset with default hyperparameters.

CLIP-ReID is a re-identification method that leverages pre-trained vision-language models like CLIP by introducing a two-stage training strategy \citep{li2023clip}. In the first stage, it generates text descriptions for each identity using learnable text tokens optimized to align with CLIP's image features. The second stage fine-tunes the image encoder with these descriptions and common ReID losses, achieving state-of-the-art performance on various datasets. On the other hand, SOLIDER focuses on self-supervised learning by generating pseudo semantic labels for human image regions (e.g., upper body, shoes) and using a semantic classification task to learn rich semantic information \citep{chen2023beyond}. It also introduces a semantic controller, allowing customizable semantic and appearance information, making it versatile for diverse human-centric tasks, including ReID.

\subsection{Evaluation Metrics} To ensure consistency with existing research, we used standard evaluation metrics for all experiments in our study:

\paragraph{Cumulative Match Characteristic (CMC)}
The Cumulative Match Characteristic (CMC) curve is a key metric for assessing person re-identification (ReID) performance. It calculates the probability of a correct match appearing within the top-k ranks for a given query. We report the Rank-1, Rank-5, and Rank-10 accuracies as standard measures of model performance \citep{ye2021deep}.

\paragraph{Mean Average Precision (mAP)}
Mean Average Precision (mAP) is another essential metric for evaluating ReID performance. It combines precision and recall by computing the average precision (AP) for each query and then averaging these values across all queries. AP is the area under the precision-recall curve, reflecting how effectively the model ranks relevant results. mAP serves as a comprehensive metric that evaluates both the accuracy and completeness of retrieval, making it a widely used benchmark for comparing ReID models \citep{ye2021deep}.
\section{Results}
\label{sec:results}

This section presents the evaluation of our proposed dataset for person re-identification (ReID) tasks. We outline the dataset preparation process, describe the training and evaluation methodologies, and provide performance metrics to assess the effectiveness of the models and dataset. Detailed results include baseline evaluations, sequence-based re-identification, and various ablation studies such as cross-dataset performance, gender-based analysis, visibility-based evaluations, and facial feature analysis.

\subsection{Dataset Preparation}
The dataset is temporally divided into training and testing subsets, with the first 75\% of the annotated video footage from all cameras allocated to the training subset and the remaining 25\% reserved for testing. This division is based on video duration rather than the number of images or identities. A separate query subset is created by sampling identities from the testing set. To ensure robust evaluation, images corresponding to the query subset are excluded from the gallery during testing. The training set is utilized to fine-tune pre-trained ReID models, while the test set is dedicated solely to evaluating model performance.

Table~\ref{tab:dataset_splits} summarizes the statistics of the training, testing and query subsets, including the number of identities, images, and cameras.

\begin{table}[ht]
    \centering
    \caption{Dataset statistics for training, testing and query subsets.}
    \label{tab:dataset_splits}
    \renewcommand{\arraystretch}{1.5}
    \begin{tabular}{|l|c|c|c|} \hline
        \textbf{Subset} & \textbf{Number of IDs} & \textbf{Number of Images} & \textbf{Number of Cameras} \\
        \hline
        Train   & 1,193   & 72,393  & 17 \\
        Test    & 654     & 45,062  & 17 \\
        Query   & 654     & 1,428   & 17 \\
        \hline
    \end{tabular}
\end{table}

\subsection{Standard Fine-tuning}
A common approach for evaluating models on datasets involves fine-tuning the models on the training set and assessing their performance on the test set. This setup provides a baseline for comparison and highlights the dataset's effectiveness in training re-identification models. The results of this evaluation are presented in Table \ref{tab:main-results}.

\begin{table*}[ht]
\centering
\caption{Model performance on our dataset compared to other benchmarks.}
\label{tab:main-results}
\renewcommand{\arraystretch}{1.5}
\begin{tabular}{|l|l|c|c|c|c|} \hline
\textbf{Model} & \textbf{Dataset} & \textbf{Rank-1} & \textbf{Rank-5} & \textbf{Rank-10} & \textbf{mAP} \\
\hline
\multirow{5}{*}{SOLIDER \citep{chen2023beyond}} & IUST\_PersonReId & 51.08\% & 65.32\% & 70.76\% & 42.35\% \\
 & Market1501 & 96.55\% & 98.78\% & 99.25\% & 93.10\% \\
 & MSMT17 & 80.35\% & 88.34\% & 90.39\% & 65.36\% \\
 & Duke & 70.37\% & 82.85\% & 86.44\% & 52.89\% \\
 & DESTRE-P & 84.3\% & 86.30\% & 86.90\% & 43.60\% \\
\hline
\multirow{5}{*}{CLIP-ReID \citep{li2023clip}} & IUST\_PersonReId & 61.40\% & 72.50\% & 77.50\% & 51.60\% \\
 & Market1501 & 95.22\% & 98.28\% & 98.96\% & 89.67\% \\
 & MSMT17 & 88.82\% & 94.42\% & 95.76\% & 73.32\% \\
 & Duke & 90.57\% & 95.83\% & 97.13\% & 82.77\% \\
 & DESTRE-P & 86.93\% & 91.50\% & 92.16\% & 43.00\% \\
\hline
\end{tabular}

\end{table*}

The results reveal an average performance drop of approximately 30\% in mAP for state-of-the-art person re-identification models on the IUST\_PersonReId dataset compared to established benchmarks like Market1501 and MSMT17. This decline demonstrates the challenges introduced by the new domain, where individuals wear modest attire characteristic of Iranian culture. The increased occlusion and lack of distinct visual features make re-identification more difficult, emphasizing the need for further advancements in model robustness to handle diverse and challenging scenarios.

\subsection{Sequence-based Re-identification}
To simulate real-world scenarios where multiple images of the same individual are available, we employed a sequence-based approach. In the first step, multiple images of the same individual were selected as queries. In the second step, a majority voting mechanism was applied to the matching scores between query images and gallery identities to determine the best match.

As shown in Table \ref{tab:sequence-results}, the sequence-based approach significantly improves performance for most metrics, including Rank-5, Rank-10, and mAP. However, for Rank-1 accuracy, the improvement is less pronounced, and for the CLIP-ReID model, it even shows a slight decrease. This highlights that while temporal context is highly beneficial for reducing ambiguity and enhancing robustness, it may not always improve the top-ranked prediction.

This method is also very effective in handling challenges like changes in lighting and unfavorable angles of individuals relative to the camera. By using multiple images taken under different conditions and viewpoints, it reduces the impact of unclear or less useful frames. This makes it a strong solution for cases where single-image re-identification methods face difficulties.

\begin{table*}[ht]
\centering
\caption{Impact of sequence-based re-identification on IUST\_PersonReId performance.}
\label{tab:sequence-results}
\renewcommand{\arraystretch}{1.5}
\begin{tabular}{|l|l|c|c|c|c|} \hline
\textbf{Model} & \textbf{Dataset} & \textbf{Rank-1} & \textbf{Rank-5} & \textbf{Rank-10} & \textbf{mAP} \\
\hline
SOLIDER \citep{chen2023beyond} & IUST\_PersonReId & 55.4\% & 95.2\% & 97.5\% & 69.6\% \\
\hline
CLIP-ReID \citep{li2023clip} & IUST\_PersonReId & 57.4\% & 96.2\% & 98\% & 71\% \\
\hline
\end{tabular}
\end{table*}

\subsection{Ablation Studies}
\subsubsection{Cross-Dataset Performance Comparison}
To further investigate the generalizability of our dataset, we conduct a cross-dataset performance analysis. In this experiment, models are trained on one dataset and tested on another, enabling us to evaluate how well the learned representations transfer across different datasets. The results of this ablation study provide insights into the compatibility and domain gap between our proposed dataset and existing benchmarks. Table~\ref{tab:cross-dataset-map}, \ref{tab:cross-dataset-rank} summarizes the results of each model, where each row corresponds to the dataset used for training, and each column indicates the dataset used for testing.

\begin{table*}[ht]
    \centering
    \caption{Cross-dataset performance comparison (mAP\%). The numbers in parentheses represent the amount of performance drop relative to the diagonal element of the testing dataset. Rows represent the training dataset, while columns represent the testing dataset.}
    \label{tab:cross-dataset-map}
    \renewcommand{\arraystretch}{1.5}
    \resizebox{\textwidth}{!}{%
        \begin{tabular}{|c|l|c|c|c|c|c|}
            \hline
            \textbf{Model} & \textbf{Train / Test}   & Market1501 & Duke & MSMT17 & DESTRE-P & IUST\_PersonReId \\
            \hline
            \multirow{5}{*}{SOLIDER \citep{chen2023beyond}} 
            & Market1501              & 93.10      & 52.90 (\(\uparrow0.01\)) & 16.95 (\(\downarrow48.41\)) & 31.00 (\(\downarrow12.60\)) & 5.01 (\(\downarrow37.34\)) \\
            & Duke                    & 14.21 (\(\downarrow78.89\)) & 52.89      & 1.12 (\(\downarrow64.24\)) & 12.10 (\(\downarrow31.50\)) & 0.71 (\(\downarrow41.64\)) \\
            & MSMT17                  & 50.05 (\(\downarrow43.05\)) & 51.71 (\(\downarrow1.18\)) & 65.36 & 33.80 (\(\downarrow9.80\)) & 8.12 (\(\downarrow34.23\)) \\
            & DESTRE-P                & 23.60 (\(\downarrow69.50\)) & 22.40 (\(\downarrow30.49\)) & 7.30 (\(\downarrow58.06\)) & 43.60 & 3.30 (\(\downarrow39.05\)) \\
            & IUST\_PersonReId        & 30.15 (\(\downarrow59.52\)) & 31.06 (\(\downarrow21.83\)) & 9.30 (\(\downarrow56.06\)) & 29.00 (\(\downarrow14.60\)) & 42.35 \\
            \hline
            \multirow{5}{*}{CLIP-ReID \citep{li2023clip}} 
            & Market1501              & 89.67      & 51.09 (\(\downarrow31.68\)) & 23.52 (\(\downarrow49.80\)) & 34.40 (\(\downarrow8.60\)) & 6.69 (\(\downarrow44.89\)) \\
            & Duke                    & 41.73 (\(\downarrow47.94\)) & 82.77      & 21.72 (\(\downarrow51.60\)) & 33.76 (\(\downarrow9.24\)) & 5.77 (\(\downarrow45.81\)) \\
            & MSMT17                  & 46.12 (\(\downarrow43.55\)) & 57.60 (\(\downarrow25.17\)) & 73.32 & 36.21 (\(\downarrow6.79\)) & 13.24 (\(\downarrow38.34\)) \\
            & DESTRE-P                & 22.04 (\(\downarrow67.63\)) & 20.00 (\(\downarrow62.77\)) & 6.18 (\(\downarrow67.14\)) & 43.00 & 3.62 (\(\downarrow47.96\)) \\
            & IUST\_PersonReId        & 39.63 (\(\downarrow49.04\)) & 42.92 (\(\downarrow39.85\)) & 19.24 (\(\downarrow54.08\)) & 33.95 (\(\downarrow9.05\)) & 51.58 \\
            \hline
        \end{tabular}
    }
\end{table*}

\begin{table*}[ht]
    \centering
    \caption{Cross-dataset performance comparison (Rank-1\%). The numbers in parentheses represent the amount of performance drop relative to the diagonal element of the testing dataset. Rows represent the training dataset, while columns represent the testing dataset.}
    \label{tab:cross-dataset-rank}
    \renewcommand{\arraystretch}{1.5}
    \resizebox{\textwidth}{!}{%
        \begin{tabular}{|c|l|c|c|c|c|c|}
            \hline
            \textbf{Model} & \textbf{Train / Test}   & Market1501 & Duke & MSMT17 & DESTRE-P & IUST\_PersonReId \\
            \hline
            \multirow{5}{*}{SOLIDER \citep{chen2023beyond}} 
            & Market1501              & 96.55      & 70.19 (\(\downarrow0.18\)) & 34.71 (\(\downarrow45.64\)) & 73.90 (\(\downarrow10.40\)) & 8.80 (\(\downarrow42.28\)) \\
            & Duke                    & 32.51 (\(\downarrow64.04\)) & 70.37      & 3.41 (\(\downarrow76.94\)) & 43.80 (\(\downarrow40.50\)) & 1.30 (\(\downarrow49.78\)) \\
            & MSMT17                  & 74.82 (\(\downarrow21.73\)) & 69.30 (\(\downarrow1.07\)) & 80.35 & 77.80 (\(\downarrow6.50\)) & 12.39 (\(\downarrow38.69\)) \\
            & DESTRE-P                & 50.40 (\(\downarrow45.15\)) & 41.20 (\(\downarrow29.17\)) & 23.80 (\(\downarrow56.55\)) & 84.30 & 6.10 (\(\downarrow45.31\)) \\
            & IUST\_PersonReId        & 55.73 (\(\downarrow40.82\)) & 49.82 (\(\downarrow20.75\)) & 20.93 (\(\downarrow59.42\)) & 71.20 (\(\downarrow13.10\)) & 51.08 \\
            \hline
            \multirow{5}{*}{CLIP-ReID \citep{li2023clip}} 
            & Market1501              & 95.22      & 69.25 (\(\downarrow21.32\)) & 50.13 (\(\downarrow38.69\)) & 73.86 (\(\downarrow13.07\)) & 12.11 (\(\downarrow49.47\)) \\
            & Duke                    & 65.41 (\(\downarrow29.81\)) & 90.57      & 48.06 (\(\downarrow40.76\)) & 77.12 (\(\downarrow9.81\)) & 10.64 (\(\downarrow50.94\)) \\
            & MSMT17                  & 65.32 (\(\downarrow29.90\)) & 74.15 (\(\downarrow16.42\)) & 88.82      & 80.39 (\(\downarrow6.54\)) & 19.05 (\(\downarrow42.36\)) \\
            & DESTRE-P                & 47.98 (\(\downarrow47.24\)) & 40.98 (\(\downarrow49.59\)) & 19.99 (\(\downarrow68.83\)) & 86.93 & 6.44 (\(\downarrow44.97\)) \\
            & IUST\_PersonReId        & 63.24 (\(\downarrow32.17\)) & 61.94 (\(\downarrow28.63\)) & 41.97 (\(\downarrow46.85\)) & 78.43 (\(\downarrow8.50\)) & 61.41 \\
            \hline
        \end{tabular}
    }
\end{table*}

The results show that there is a notable drop in accuracy when models are tested across different datasets, such as from Market1501 to MSMT17 or from Duke to Market1501, indicating significant domain gaps even among existing benchmarks. However, the performance drop is even more pronounced when testing on our IUST\_PersonReId dataset, demonstrating the unique challenges it presents.

\subsubsection{Gender-based Query and Gallery Setup} 
To explore model performance in gender-specific scenarios, we tested the following configurations:  
\begin{itemize}
    \item Female Queries: Queries consisting only of female images, with the gallery containing both male and female identities.  
    \item Male Queries: Queries consisting only of male images, with the gallery containing both male and female identities.  
\end{itemize}

This setup examines how well models differentiate gender-based features. Initially, the models were trained on the original dataset, which is imbalanced, with a majority of male identities (1,338) compared to female identities (509). To investigate the effect of this imbalance, we conducted an additional experiment where the dataset was balanced by undersampling male identities to match the number of female identities. The models were then fine-tuned on the balanced dataset. Results from both the original and balanced datasets are presented in Table \ref{tab:gender-balanced-comparison}.  



\begin{table*}[ht]
\centering
\caption{Gender-wise Performance Comparison Between Original and Balanced Datasets for Each Model}
\label{tab:gender-balanced-comparison}
\renewcommand{\arraystretch}{1.5}
\begin{tabular}{|l|l|cc|cc|cc|cc|}
\hline
\textbf{Gender} & \textbf{Model} 
& \multicolumn{2}{c|}{\textbf{Rank-1}} 
& \multicolumn{2}{c|}{\textbf{Rank-5}} 
& \multicolumn{2}{c|}{\textbf{Rank-10}} 
& \multicolumn{2}{c|}{\textbf{mAP}} \\
\cline{3-10}
 & 
 & \textbf{Orig.} & \textbf{Bal.} 
 & \textbf{Orig.} & \textbf{Bal.} 
 & \textbf{Orig.} & \textbf{Bal.} 
 & \textbf{Orig.} & \textbf{Bal.} \\
\hline
\multirow{2}{*}{Male} 
& SOLIDER \citep{chen2023beyond} & 59.79\% & 40.44\% & 73.03\% & 43.56\% & 78.05\% & 48.00\% & 49.20\% & 38.42\% \\
& CLIP-ReID \citep{li2023clip}    & 68.30\% & 69.57\% & 78.90\% & 82.34\% & 82.80\% & 84.78\% & 62.21\% & 64.45\% \\
\hline
\multirow{2}{*}{Female} 
& SOLIDER \citep{chen2023beyond} & 34.98\% & 30.34\% & 51.08\% & 37.15\% & 57.27\% & 40.87\% & 29.67\% & 24.98\% \\
& CLIP-ReID \citep{li2023clip}    & 45.60\% & 44.68\% & 57.90\% & 58.80\% & 65.30\% & 65.74\% & 36.27\% & 35.01\% \\
\hline
\end{tabular}
\end{table*}

From the results, we observe that the re-identification of females remains significantly more challenging than males, even with a balanced dataset. While balancing mitigates the dataset's bias, it does not eliminate the intrinsic challenges associated with identifying females, such as the limited distinctive features and similarity in appearance due to hijabs and modest attire. This highlights the need for models that can better handle these cultural and clothing-specific challenges.

\subsubsection{Impact of Visibility on Re-identification Performance} 
The dataset was categorized into three visibility levels—clear, partial, and occluded—based on the visibility of 18 keypoints extracted for each person using the AlphaPose \citep{fang2022alphapose} pose estimation model. The categorization criteria are as follows:
\begin{itemize}
    \item \textbf{Clear:} Most of the face and upper body are visible, with no more than 4 missing keypoints.
    \item \textbf{Partial:} At least 9 out of 18 keypoints are visible.
    \item \textbf{Occluded:} Fewer than 9 out of 18 keypoints are visible.
\end{itemize}

\begin{figure}[ht]
    \centering
    \begin{minipage}{0.48\textwidth}
        \centering
        \includegraphics[width=\textwidth]{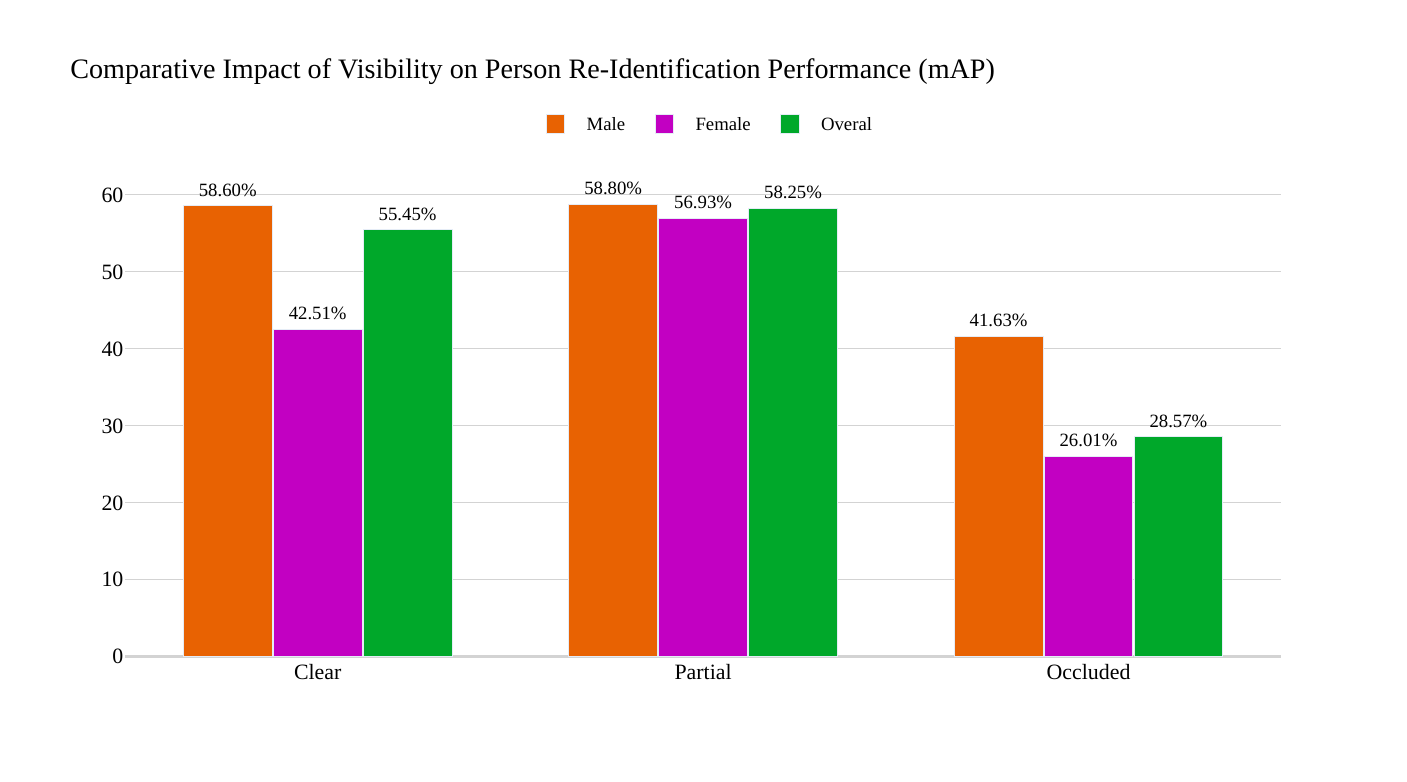}
        \caption{Mean Average Precision of CLIP-ReID Across Visibility Categories by Gender.}
        \label{fig:visibility}
    \end{minipage}
\end{figure}

These categories were determined based on the visibility and clarity of keypoints, reflecting variations in occlusion, pose, or camera angle. Mean average precision was computed separately for each visibility category and gender to evaluate the model’s performance under varying visual conditions. As shown in Figure~\ref{fig:visibility}, the model achieves the highest mAP on clear examples, with a noticeable performance drop under partial and occluded settings. 


\begin{table}[ht]
    \centering
    \caption{Number of Male and Female Queries per Visibility Category.}
    \label{tab:visibility-gender-distribution}
    \renewcommand{\arraystretch}{1.25}
    \begin{tabular}{|l|c|c|} \hline
        \textbf{Category} & \textbf{\#Male} & \textbf{\#Female} \\
        \hline
        Clear   & 401   & 83 \\
        Partial    & 58     & 21 \\
        Occluded   & 17     & 74 \\
        \hline
    \end{tabular}
\end{table}



Interestingly, the gender-wise breakdown reveals a performance disparity linked to the distribution of identities across visibility categories. Table~\ref{tab:visibility-gender-distribution} shows that clear queries are predominantly male, who are generally more visible in the dataset. In contrast, the majority of occluded queries are female, often due to cultural attire such as the hijab, which contributes to increased occlusion and reduced keypoint visibility. This skew in data distribution helps explain why female performance is lower under the clear setting but surpasses male performance under partial visibility, where fewer clear cues are available for either gender. Overall, these findings highlight the critical role of visibility and gender-specific occlusion in the re-identification task.

\subsubsection{Facial Feature Analysis}
\begin{figure}[h]
    \centering
    \begin{minipage}{0.48\textwidth}
        \centering
        \includegraphics[width=\textwidth]{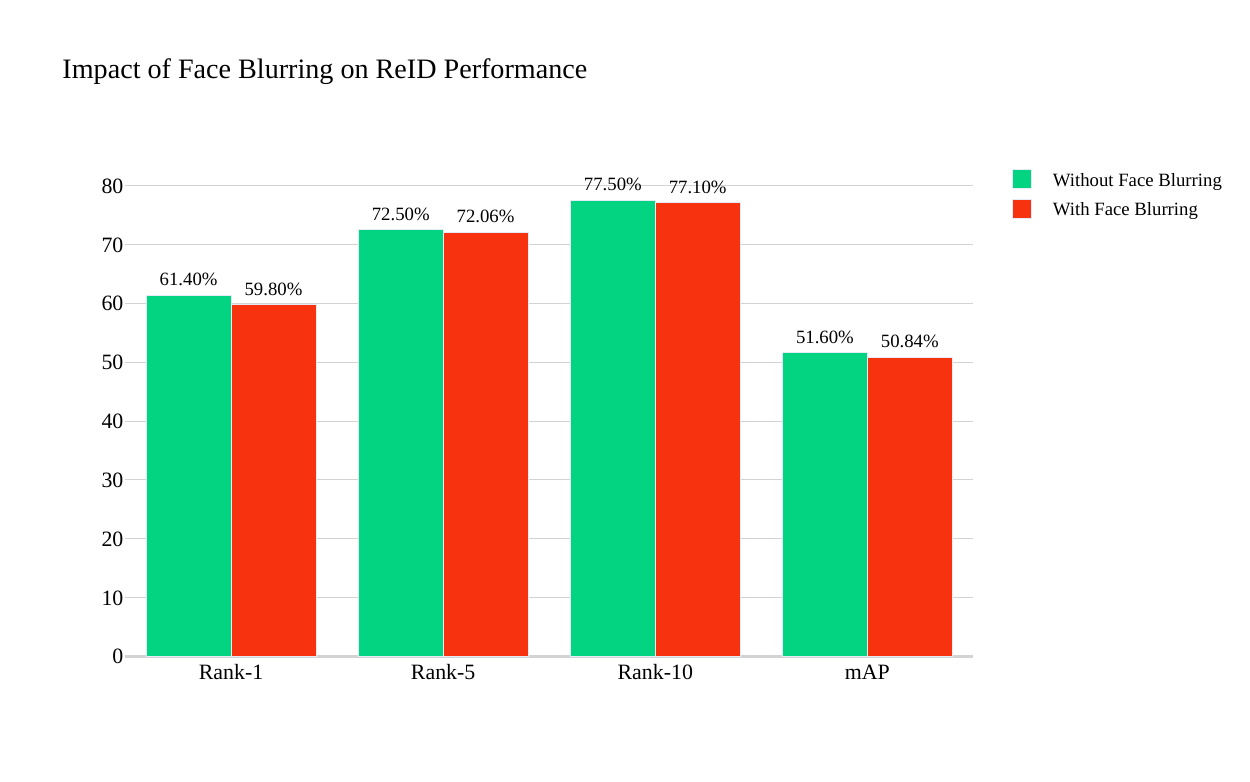}
        \caption{Impact of face blurring on the person re-identification performance.}
        \label{fig:blur_impact}
    \end{minipage}
    \hfill 
    \begin{minipage}{0.48\textwidth}
        \centering
        \includegraphics[width=\textwidth]{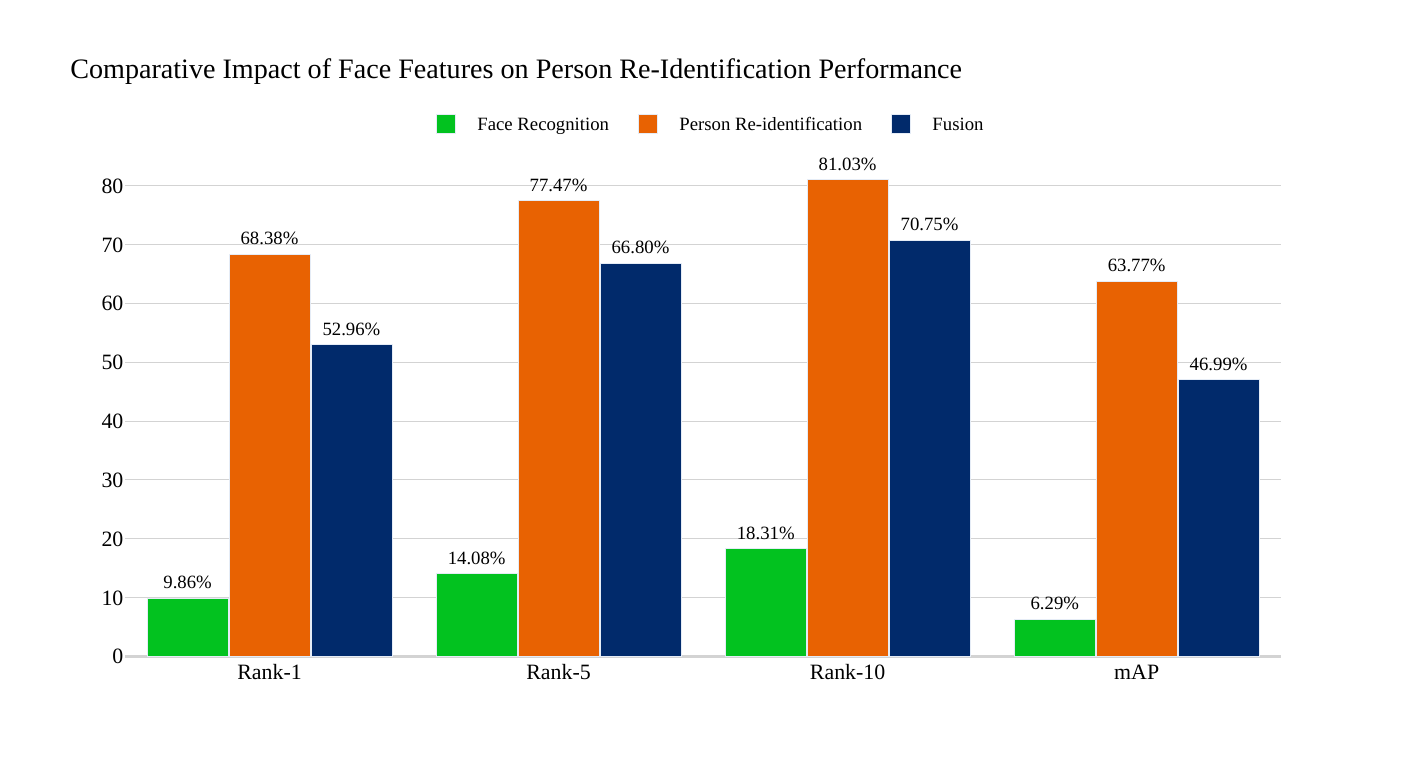}
        \caption{Evaluation of face recognition, person re-identification, and their combined fusion performance.}
        \label{fig:face_recognition}
    \end{minipage}
\end{figure}

To investigate the role of facial features in person re-identification, we conducted a series of experiments focused on face visibility and its potential integration with full-body cues. First, we examined the impact of face blurring, which was applied to the publicly released version of our dataset for privacy and ethical considerations. As shown in Figure~\ref{fig:blur_impact}, the results indicate that blurring faces leads to only a marginal reduction in performance, suggesting that facial features, while informative in certain cases, are not the dominant cues leveraged by person ReID models in real-world surveillance scenarios.

Next, we created a subset of the test set where the face was clearly visible in the query image and at least 10\% of the corresponding gallery images. We performed three separate experiments on this subset: (1) using  Additive Angular Margin Loss (ArcFace \citep{deng2019arcface}), a face recognition model to extract facial embeddings and perform matching based on Euclidean distance, (2) using a full-body person re-identification model (CLIP-ReID) with the same matching protocol, and (3) applying a fusion strategy that combined the distance matrices from both models to jointly leverage facial and body cues. As illustrated in Figure~\ref{fig:face_recognition}, the person re-identification model outperformed the face-only approach, and the fusion of face and body distances slightly degraded the overall accuracy. This degradation is attributed to the limitations inherent in surveillance footage, such as low face resolution, suboptimal lighting, and non-frontal angles, which reduce the reliability of facial features.

These findings emphasize that in realistic, in-the-wild settings, especially those involving modest attire and partially occluded faces robust person re-identification is best achieved through comprehensive body-based feature representations rather than relying heavily on facial cues.

\section{Conclusion}
\label{sec:conclusion}

This paper introduces the IUST\_PersonReId dataset, designed to address the unique challenges of person re-identification in Islamic countries, where modest clothing, particularly hijabs, is prevalent. The dataset captures diverse scenes from Iran and Iraq, with variations in lighting, camera angles, and seasonal clothing. Through a multi-step annotation process involving automated tracking and manual refinement, we ensure high-quality frames for each individual.

Our experiments with state-of-the-art models, such as SOLIDER and CLIP-ReID, demonstrate a significant performance drop when tested on IUST\_PersonReId compared to traditional benchmarks like Market1501 and MSMT17. This highlights the difficulties posed by modest attire, especially for women, where occlusion and limited distinctive features impact re-identification accuracy. Sequence-based evaluation, which utilizes multiple images of an individual, shows promising improvements by leveraging temporal context.

Further analysis confirms a considerable domain gap between IUST\_PersonReId and existing datasets, emphasizing the need for datasets that better represent specific cultural contexts. Gender-based performance disparities were also observed, with female re-identification being particularly challenging due to the similarities in appearance caused by hijab usage. Additionally, visibility-based evaluation underscores the importance of clear and visible body keypoints for accurate re-identification.

The IUST\_PersonReId dataset is a valuable resource for developing models that are more robust to modest attire and cultural clothing variations. It provides an essential foundation for advancing person re-identification technologies that are applicable in diverse cultural contexts and ensure fairness in real-world applications. Future research should focus on expanding the dataset, addressing the challenges posed by modest attire, and exploring bias mitigation strategies to improve re-identification in such contexts.

    
    

\appendix


 \bibliographystyle{elsarticle-num} 
 \bibliography{cas-refs}

\begin{thebibliography}{10}
\expandafter\ifx\csname url\endcsname\relax
  \def\url#1{\texttt{#1}}\fi
\expandafter\ifx\csname urlprefix\endcsname\relax\def\urlprefix{URL }\fi
\expandafter\ifx\csname href\endcsname\relax
  \def\href#1#2{#2} \def\path#1{#1}\fi

\bibitem{perkowitz2021bias}
S.~Perkowitz, The {Bias} in the {Machine}: Facial {Recognition} {Technology} and {Racial} {Disparities}, MIT Case Studies in Social and Ethical Responsibilities of Computing~(Winter 2021), https://mit-serc.pubpub.org/pub/bias-in-machine (feb 5 2021).

\bibitem{li2014deepreid}
W.~Li, R.~Zhao, T.~Xiao, X.~Wang, Deepreid: Deep filter pairing neural network for person re-identification, IEEE Conference on Computer Vision and Pattern Recognition (2014) 152--159.

\bibitem{narayan2017person}
N.~Narayan, N.~Sankaran, D.~Arpit, K.~Dantu, S.~Setlur, V.~Govindaraju, Person re-identification for improved multi-person multi-camera tracking by continuous entity association, in: Proceedings of the IEEE Conference on Computer Vision and Pattern Recognition Workshops, 2017, pp. 64--70.

\bibitem{zheng2017unlabeled}
Z.~Zheng, L.~Zheng, Y.~Yang, Unlabeled samples generated by gan improve the person re-identification baseline in vitro, in: Proceedings of the IEEE international conference on computer vision, 2017, pp. 3754--3762.

\bibitem{zheng2015scalable}
L.~Zheng, L.~Shen, L.~Tian, S.~Wang, J.~Wang, Q.~Tian, Scalable person re-identification: A benchmark, IEEE Transactions on Image Processing 24~(11) (2015) 4177--4187.

\bibitem{wei2018person}
L.~Wei, S.~Zhang, W.~Gao, Q.~Tian, Person transfer gan to bridge domain gap for person re-identification, IEEE Conference on Computer Vision and Pattern Recognition (2018) 79--88.

\bibitem{fu2021unsupervised}
D.~Fu, D.~Chen, J.~Bao, H.~Yang, L.~Yuan, L.~Zhang, H.~Li, D.~Chen, Unsupervised pre-training for person re-identification, in: Proceedings of the IEEE/CVF conference on computer vision and pattern recognition, 2021, pp. 14750--14759.

\bibitem{fu2022large}
D.~Fu, D.~Chen, H.~Yang, J.~Bao, L.~Yuan, L.~Zhang, H.~Li, F.~Wen, D.~Chen, Large-scale pre-training for person re-identification with noisy labels, in: Proceedings of the IEEE/CVF conference on computer vision and pattern recognition, 2022, pp. 2476--2486.

\bibitem{felzenszwalb2009object}
P.~F. Felzenszwalb, R.~B. Girshick, D.~McAllester, D.~Ramanan, Object detection with discriminatively trained part-based models, IEEE transactions on pattern analysis and machine intelligence 32~(9) (2009) 1627--1645.

\bibitem{zheng2016mars}
L.~Zheng, Z.~Bie, Y.~Sun, J.~Wang, C.~Su, S.~Wang, Q.~Tian, Mars: A video benchmark for large-scale person re-identification, European Conference on Computer Vision (2016) 868--884.

\bibitem{dehghan2015gmmcp}
A.~Dehghan, S.~Modiri~Assari, M.~Shah, Gmmcp tracker: Globally optimal generalized maximum multi clique problem for multiple object tracking, in: Proceedings of the IEEE conference on computer vision and pattern recognition, 2015, pp. 4091--4099.

\bibitem{ristani2016performance}
E.~Ristani, F.~Solera, R.~Zou, R.~Cucchiara, C.~Tomasi, Performance measures and a data set for multi-target, multi-camera tracking, in: European conference on computer vision, Springer, 2016, pp. 17--35.

\bibitem{benenson2015ten}
R.~Benenson, M.~Omran, J.~Hosang, B.~Schiele, Ten years of pedestrian detection, what have we learned?, in: Computer Vision-ECCV 2014 Workshops: Zurich, Switzerland, September 6-7 and 12, 2014, Proceedings, Part II 13, Springer, 2015, pp. 613--627.

\bibitem{wu2018exploit}
Y.~Wu, Y.~Lin, X.~Dong, Y.~Yan, W.~Ouyang, X.~Yang, Exploit the unknown gradually: One-shot video-based person re-identification by stepwise learning, IEEE Conference on Computer Vision and Pattern Recognition (2018) 5177--5186.

\bibitem{gou2018systematic}
M.~Gou, Z.~Wu, A.~Rates-Borras, O.~Camps, R.~J. Radke, et~al., A systematic evaluation and benchmark for person re-identification: Features, metrics, and datasets, IEEE transactions on pattern analysis and machine intelligence 41~(3) (2018) 523--536.

\bibitem{felsen2017will}
P.~Felsen, Y.-C. Tsai, W.-C. Huang, Y.-C. Huang, Will it collide? probabilistic trajectory prediction for sports video analysis, IEEE Conference on Computer Vision and Pattern Recognition (2017) 3999--4008.

\bibitem{zheng2018rpifield}
M.~Zheng, S.~Karanam, R.~J. Radke, Rpifield: A new dataset for temporally evaluating person re-identification, in: Proceedings of the IEEE conference on computer vision and pattern recognition workshops, 2018, pp. 1893--1895.

\bibitem{giancola2022soccernet}
S.~Giancola, A.~Cioppa, A.~Deli{\`e}ge, F.~Magera, V.~Somers, L.~Kang, X.~Zhou, O.~Barnich, C.~De~Vleeschouwer, A.~Alahi, et~al., Soccernet 2022 challenges results, in: Proceedings of the 5th International ACM Workshop on Multimedia Content Analysis in Sports, 2022, pp. 75--86.

\bibitem{wu2020destre}
Y.~Wu, Y.~Lin, X.~Dong, Y.~Yan, W.~Ouyang, X.~Yang, Destre: A dataset for person re-identification in open-world surveillance, IEEE Transactions on Pattern Analysis and Machine Intelligence (2020).

\bibitem{shu2021large}
X.~Shu, X.~Wang, X.~Zang, S.~Zhang, Y.~Chen, G.~Li, Q.~Tian, Large-scale spatio-temporal person re-identification: Algorithms and benchmark, IEEE Transactions on Circuits and Systems for Video Technology 32~(7) (2021) 4390--4403.

\bibitem{song2018region}
G.~Song, B.~Leng, Y.~Liu, C.~Hetang, S.~Cai, Region-based quality estimation network for large-scale person re-identification, IEEE Conference on Computer Vision and Pattern Recognition (2018) 6199--6208.

\bibitem{yang2023towards}
S.~Yang, Y.~Zhou, Z.~Zheng, Y.~Wang, L.~Zhu, Y.~Wu, Towards unified text-based person retrieval: A large-scale multi-attribute and language search benchmark, in: Proceedings of the 31st ACM International Conference on Multimedia, 2023, pp. 4492--4501.

\bibitem{Jocher_YOLOv5_by_Ultralytics_2020}
G.~Jocher, \href{https://github.com/ultralytics/yolov5}{{YOLOv5 by Ultralytics}} (May 2020).
\newblock \href {https://doi.org/10.5281/zenodo.3908559} {\path{doi:10.5281/zenodo.3908559}}.
\newline\urlprefix\url{https://github.com/ultralytics/yolov5}

\bibitem{zhang2022bytetrack}
Y.~Zhang, P.~Sun, Y.~Jiang, D.~Yu, F.~Weng, Z.~Yuan, P.~Luo, W.~Liu, X.~Wang, Bytetrack: Multi-object tracking by associating every detection box, in: European conference on computer vision, Springer, 2022, pp. 1--21.

\bibitem{zhang2021fairmot}
Y.~Zhang, C.~Wang, X.~Wang, W.~Zeng, W.~Liu, Fairmot: On the fairness of detection and re-identification in multiple object tracking, International journal of computer vision 129 (2021) 3069--3087.

\bibitem{ppdet2019}
P.~Authors, Paddledetection, object detection and instance segmentation toolkit based on paddlepaddle., \url{https://github.com/PaddlePaddle/PaddleDetection} (2019).

\bibitem{CVAT_ai_Corporation_Computer_Vision_Annotation_2023}
{CVAT.ai Corporation}, \href{https://github.com/cvat-ai/cvat}{{Computer Vision Annotation Tool (CVAT)}} (Nov. 2023).
\newline\urlprefix\url{https://github.com/cvat-ai/cvat}

\bibitem{ROSEReID}
N.~T.~U. ROSE~Lab, Person re-identification, \url{https://www.ntu.edu.sg/rose/research-focus/deep-learning-video-analytics/person-re-identification}.

\bibitem{zhao2018semi}
T.~Zhao, S.~Liao, Z.~Lei, Semi-automatic data annotation tool for person re-identification across multi cameras, in: 2018 IEEE International Conference on Big Data (Big Data), IEEE, 2018, pp. 4672--4677.

\bibitem{mittal2012no}
A.~Mittal, A.~K. Moorthy, A.~C. Bovik, No-reference image quality assessment in the spatial domain, IEEE Transactions on image processing 21~(12) (2012) 4695--4708.

\bibitem{chen2023beyond}
W.~Chen, X.~Xu, J.~Jia, H.~Luo, Y.~Wang, F.~Wang, R.~Jin, X.~Sun, Beyond appearance: a semantic controllable self-supervised learning framework for human-centric visual tasks, in: Proceedings of the IEEE/CVF conference on computer vision and pattern recognition, 2023, pp. 15050--15061.

\bibitem{li2023clip}
S.~Li, L.~Sun, Q.~Li, Clip-reid: exploiting vision-language model for image re-identification without concrete text labels, in: Proceedings of the AAAI Conference on Artificial Intelligence, Vol.~37, 2023, pp. 1405--1413.

\bibitem{radford2021learning}
A.~Radford, J.~W. Kim, C.~Hallacy, A.~Ramesh, G.~Goh, S.~Agarwal, G.~Sastry, A.~Askell, P.~Mishkin, J.~Clark, et~al., Learning transferable visual models from natural language supervision, in: International conference on machine learning, PMLR, 2021, pp. 8748--8763.

\bibitem{ye2021deep}
M.~Ye, J.~Shen, G.~Lin, T.~Xiang, L.~Shao, S.~C. Hoi, Deep learning for person re-identification: A survey and outlook, IEEE transactions on pattern analysis and machine intelligence 44~(6) (2021) 2872--2893.

\bibitem{fang2022alphapose}
H.-S. Fang, J.~Li, H.~Tang, C.~Xu, H.~Zhu, Y.~Xiu, Y.-L. Li, C.~Lu, Alphapose: Whole-body regional multi-person pose estimation and tracking in real-time, IEEE Transactions on Pattern Analysis and Machine Intelligence 45~(6) (2022) 7157--7173.

\bibitem{deng2019arcface}
J.~Deng, J.~Guo, N.~Xue, S.~Zafeiriou, Arcface: Additive angular margin loss for deep face recognition, in: Proceedings of the IEEE/CVF Conference on Computer Vision and Pattern Recognition, 2019, pp. 4690--4699.

\end{thebibliography}





\end{document}